\documentclass{article}

\usepackage[final]{arXiv} 

\usepackage[utf8]{inputenc} 
\usepackage[T1]{fontenc}    
\usepackage{hyperref}       
\usepackage{url}            
\usepackage{booktabs}       
\usepackage{amsfonts}       
\usepackage{nicefrac}       
\usepackage{microtype}      

\usepackage[english]{babel}

\usepackage{graphicx}
\usepackage{animate}
\usepackage{epsfig} 				
\usepackage{epstopdf}

\usepackage{listings}
\usepackage{color}
\usepackage{nameref}
\usepackage{hyperref}
\usepackage[colorinlistoftodos]{todonotes}
\usepackage{amsmath}	 			
\usepackage{amssymb}  				

\usepackage{dsfont}			
\usepackage{mathtools}

\usepackage{epigraph}
\usepackage{lscape}
\usepackage[]{nomencl}				
\usepackage{algorithm}
\usepackage{algorithmic}
\usepackage{multicol}
\usepackage{multirow}
\usepackage{etoolbox}

\usepackage{caption}
\usepackage{subcaption}

\usepackage{wrapfig}
\usepackage{multimedia}

\usepackage{units}
\usepackage{flushend}

\usepackage{floatflt}

\usepackage{url}

\usepackage[absolute,overlay]{textpos}

\usepackage{bibentry}

\usepackage{tikz}
\usepgflibrary{arrows}	

\usepackage{float}
\usepackage[utf8]{inputenc}
\usepackage[english]{babel}
\usepackage{graphics}
\usepackage{animate}

\usepackage{listings}

\usepackage[nostamp]{draftwatermark}

\usepackage{extarrows}		

\allowdisplaybreaks[1]

\newcommand{\playvideo}[1]{\href{run:#1}{\includegraphics[scale=0.12]{\RCPath fig/empty}}}

\usepackage{lipsum}
\usepackage{enumitem}
\usepackage{framed}

\newcommand{\email}[1]{\href{mailto:#1}{\nolinkurl{#1}}}

\newcommand{\link}[1]{\colora{\url{#1}}}

\renewcommand{\sec}[1]{Section~\ref{#1}}
\newcommand{\fig}[1]{Figure~\ref{#1}}
\newcommand{\eq}[1]{Equation~\eqref{#1}}

\newcommand{\alg}[1]{Algorithm~\ref{#1}}

\definecolor{matlab1}{rgb}{0,0,1}
\definecolor{matlab2}{rgb}{0,0.5,0}
\definecolor{matlab3}{rgb}{1,0,0}
\definecolor{matlab4}{rgb}{0,0.75,0.75}
\definecolor{matlab5}{rgb}{0.75,0,0.75}
\definecolor{matlab6}{rgb}{0.75,0.75,0}
\definecolor{matlab7}{rgb}{0.25,0.25,0.25}

\definecolor{darkgreen}{rgb}{0,0.5,0}		
\definecolor{purple}{rgb}{0.75,0,0.75}
\definecolor{pink}{rgb}{1,0.4,0.6}

\newcommand{\capitalize}[1]{\expandafter\MakeUppercase\expandafter{#1}}

\newcommand{\colora}[1]{{\usebeamercolor[fg]{framesubtitle}#1}}

\makeatletter
\newcommand*{\compress}{\@minipagetrue}
\makeatother

\renewcommand{\vec}[1]{\boldsymbol{#1}}				
\newcommand{\mat}[1]{\boldsymbol{\capitalize{#1}}}		

\newcommand{\dataset}[0]{\mathcal{D}}

\DeclareMathOperator*{\minimize}{\text{arg min}}

\newcommand{\objfuncNo}[0]{J}					

\newcommand{\respsurfNo}[0]{\hat \objfuncNo}			
\newcommand{\acqfuncNo}[0]{\alpha}				

\newcommand{\q}{\vec{q}}					
\ifdef{\dq}{\renewcommand{\dq}{\dot{\q}}}{\newcommand{\dq}{\dot{\q}}}

\newcommand{\state}[0]{s}
\newcommand{\control}[0]{a}
\newcommand{\pol}[0]{\pi}

\newcommand{\dyn}[0]{f}

\newcommand{\cost}[0]{J}
\newcommand{\runcost}[0]{l}

\newcommand{\params}[0]{\theta}
\newcommand{\gpinput}[0]{z}
\newcommand{\gpfun}[0]{g}

\newtheorem{remark}{Remark}

\title{MBMF: Model-Based Priors for \\Model-Free Reinforcement Learning}

\author{
Somil Bansal\quad Roberto Calandra\quad Kurtland Chua\quad Sergey Levine\quad Claire Tomlin\\
  Department of Electrical Engineering and Computer Sciences\\
  University of California Berkeley, United States\\
  \texttt{\{somil, roberto.calandra, kchua, svlevine, tomlin\}@berkeley.edu} \\
}

\begin{document}
\maketitle


\begin{abstract}
	Reinforcement Learning is divided in two main paradigms: model-free and model-based.
Each of these two paradigms has strengths and limitations, and has been successfully applied to real world domains that are appropriate to its corresponding strengths.
In this paper, we present a new approach aimed at bridging the gap between these two paradigms that is at the same time data-efficient and cost-savvy.
We do so by learning a probabilistic dynamics model and leveraging it as a prior for the intertwined model-free optimization.
As a result, our approach can exploit the generality and structure of the dynamics model, but is also capable of ignoring its inevitable inaccuracies, by directly incorporating the evidence provided by the direct observation of the cost.
Preliminary results demonstrate that our approach outperforms purely model-based and model-free approaches, as well as the approach of simply switching from a model-based to a model-free setting.
\end{abstract}


\section{Introduction}
	Reinforcement learning (RL) methods can generally be divided into Model-Free (MF) approaches, in which the cost is directly optimized, and Model-Based (MB) approaches, which additionally employ and/or learn a model of the environment.
Both of these approaches have different strengths and limitations~\citep{Deisenroth2013}.
Typically, MF approaches are very effective at learning complex policies, but the convergence might require millions of trials and lead to (globally sub-optimal) local minima.
On the other hand, MB approaches have the theoretical benefit of being able to better generalize to new tasks and environments, and in practice they can drastically reduce the number of trials required~\citep{Deisenroth2015,Levine2014}.
However, for this generalization, an accurate model is necessary (either engineered or learned), which on its own can be challenging to acquire.
This issue is very crucial since any bias in the model does not translate to a proportional bias in the policy -- a weakly biased model might result in a strongly biased policy.
As a consequence, MB approaches have been often limited to low-dimensional spaces, and often require a significant degree of engineering to perform well.
Thus, it is desirable to design approaches that can leverage the respective advantages and overcome the challenges of each approach.  

A second motivation for understanding and bridging the gap between MB and MF approaches is provided by neuroscience.
Evidence in neuroscience suggests that humans employ both MF and MB approaches for learning new skills, and switch between the two during the learning process~\citep{Glaescher2010}.
From the machine learning perspective, one possible explanation for this behavior can be found in the concept of bounded rationality~\citep{Simon1982}.
The switch from a MB approach to a MF approach after learning sufficiently good policies can be motivated by the need to devote the limited computational resources to new challenging tasks, while still being able to solve the previous tasks (with a sufficient generalization capability).
However, in the reinforcement learning community, there are limited coherent frameworks that combine these two approaches.

In this paper, we propose a probabilistic framework that integrates MB and MF approaches.
This bridging is achieved by considering the cost estimated by the MB component as the prior for the intertwined probabilistic MF component.
In particular, we learn a dynamics model from scratch which is used to compute the trajectory distribution corresponding to a given policy, which in turn can be used to estimate the cost of the policy. 
This estimate is used by a Bayesian Optimization-based MF policy search to guide the policy exploration.
In essence, this probabilistic framework allows us to model and combine the uncertainty in the cost estimates of the two methods.
The advantage of doing so is to exploit the structure and generality of the dynamics model throughout the entire state-action space.
At the same time, the evidence provided by the observation of the actual cost can be integrated into the estimation of the posterior.

We demonstrate our method on a 2D navigation task and a more complex simulated manipulation task that requires pushing an object to a goal position.
Our results show that the proposed approach can overcome the model bias and inaccuracies in the dynamics model to learn a well-performing policy, and yet retain and improve upon the fast convergence rate of MB approaches. 
%

\section{Related Work} \label{sec:related_work}
	Gaussian processes have been widely used in the literature to learn dynamics model~\citep{Deisenroth2015,Nguyen-Tuong2009,nguyen2011model,Schreiter2015,pan2014probabilistic} and for control purposes~\citep{Deisenroth2013,Kocijan2004,Calandra2015a}.
	By learning a forward dynamics model, it is possible to predict the sequence of states (e.g., a trajectory) generated by the given policy.
	The main hypothesis here is that the learning of the dynamics is a good proxy, full of structure, that allows to predict the behavior of a given policy without evaluating it on the real system.
	This structure is particularly valuable, as the alternative would be to directly predict the cost from the policy parameters (e.g., in Bayesian optimization (BO)), which can be challenging, especially for high dimensional policies with thousands or millions of parameters. 
	However, MB approaches do not usually incorporate evidence from the cost.
	Hence, if the model is inaccurate (e.g., due to intrinsic limitations of the models or the compounding of the inaccuracies over trajectory propagation) the expected cost might be wrong, even if the cost has been measured for the considered policy.
	This issue is often referred to as \emph{model bias}~\citep{Deisenroth2011}.
	
	To overcome the model bias, \citep{McHutchon2014} proposed to optimize not only the expected cost, but to also incorporate the predicted variance.
	This modification results in a exploration-exploitation trade-off that very closely connects to the acquisition functions used in BO.
	However, unlike our work, \citep{McHutchon2014} does not try to make use of MF approaches and therefore its approach can be considered as a specific case of our framework where, once again, the evidence derived from directly observing the cost is disregarded.   		

	Recently, it has been proposed that a solution to overcoming the model bias is to directly train the dynamics in a goal-oriented manner~\citep{Bansal2017,Donti2017} using the cost observations.
	However, this approach has the drawback that the generality of the dynamics model is lost.
	Moreover, directly optimizing the high dimensional dynamics in a goal-oriented fashion can be very challenging.
	In contrast, we learn a general dynamics model and yet take the cost observations into account which allows us to overcome the limitations of both pure MB and MF methods. 

	Several prior works have sought to combine MB and MF reinforcement learning, typically with the aim of accelerating MF learning while minimizing the effects of model bias on the final policy. 
	\citep{Gu2016} proposed a method that generates synthetic experience for MF learning using a learned model, and explored the types of models that might be suitable for continuous control tasks. 
	This work follows on a long line of research into using learned models for generating synthetic experience, including the classic Dyna framework~\citep{Sutton1991}. 
	Authors in \citep{heess2015} use models for improving the accuracy of MF value function backups. 
	In other works, models have been used to produce a good initialization for the MF component~\citep{farshidian2014,nagabandi2017}. 
	However, our method directly combines MB and MF approaches into a single RL method without using synthetic samples that degrade with the modeling errors.
	
	\citep{Chebotar2017} also proposed to combine MB and MF methods, with the MF algorithm learning the residuals from the MB return estimates. 
	Our approach also uses the MB return estimates as a bias for MF learning, but in contrast to \citep{Chebotar2017}, the MB component is incorporated as a prior mean into a Bayesian model-free update, which allows our approach to reason about the confidence of the MF estimates across the entire policy space. 
	Our approach is perhaps most similar to \citep{wilson2014} wherein a linear model is learned in the feature space which is used to provide a prior mean for the MF updates. The features used to learn the model are hand-picked. In contrast, we employ general dynamics models that are learned from scratch.


\section{Problem Formulation} 
\label{sec:formulation}
	The goal of reinforcement learning is to learn a policy that maximizes the sum of future rewards (or equivalently minimize the sum of future costs). 
In particular, we consider a discrete-time, potentially stochastic and non-linear, dynamical system
\begin{equation} \label{eqn:dyn}
\state_{k+1} = \dyn(\state_k, \control_k)\,,
\end{equation}
where $\state_k \in \mathbb{R}^n$ and $\control_k \in \mathbb{R}^m$ denote the state and the action of the system respectively at time-step $k$, and $\dyn: \mathbb{R}^n \times \mathbb{R}^m \mapsto \mathbb{R}^n$ is the state transition map.
Our objective is to find the parameters $\params$ of the policy $\pol(\state_k, \params)$ that minimizes a given cost function subject to the system dynamics.
In the finite horizon case, we aim to minimize the cost function
\begin{equation} \label{eqn:cost_fcn}
\cost^{\pol}(k, \state) = \mathbb{E}\left[\sum\limits_{i=k}^{T-1}\runcost_i(\state_i, \control_i) + \runcost_T(\state_T)\right]\,,
\end{equation}
where $T$ is time horizon, and $\runcost_i$ is the cost function at time-step~$i$. One of the key challenges in designing the policy $\pi$ is that the system dynamics $\dyn$ of \eq{eqn:dyn} are typically unknown. 
In this work, we propose a novel approach that combines MF and MB methods to learn the optimal policy $\pol(\state_k, \params^{*})$.


\section{Background} 
\label{sec:background}
	Our general approach will be to learn the dynamics model of the system, and use Bayesian Optimization (BO) to find the optimal policy parameters.
In particular, we use a Gaussian Process (GP) to model the underlying objective function in BO.
In this section, we provide a brief overview of GPs and BO.
In the next section, we combine the learned dynamics model with BO to overcome some of the challenges that pure MB and pure MF methods face.

\subsection{Gaussian Process (GP)} \label{sec:gp}
GPs are a state-of-the-art probabilistic regression method~\citep{Rasmussen2006}.
In general, a GP can be used to model a nonlinear map, $\gpfun(\gpinput): \mathbb{R}^{q_1} \rightarrow \mathbb{R}^{q_2}$, from an input vector $\gpinput$ to the function value $\gpfun(\gpinput)$. 
Hence, we assume that function values $\gpfun$, associated with different values of $\gpinput$, are random variables and that any finite number of these random variables have a joint Gaussian distribution~\citep{Rasmussen2006}.
For GPs, we define a prior mean function, $m(\gpinput)$, and a covariance function (or kernel), $k(\gpinput_i,\gpinput_j)$, which defines the covariance between any two function values, $\gpfun(\gpinput_i)$ and $\gpfun(\gpinput_j)$. 
The choice of kernel is problem-dependent and encodes general assumptions such as smoothness of the unknown function. 
In this work, we employ the squared exponential kernel where the hyperparameters are optimized by maximizing the marginal likelihood~\citep{Rasmussen2006}. 

The GP framework can be used to predict the distribution of the function~$\gpfun(\gpinput^{*})$ at an arbitrary input~$\gpinput^{*}$ based on the past observations, $\dataset=\{\gpinput_i,\gpfun(\gpinput_i)\}_{i=1}^l$. 
Conditioned on $\dataset$, the prediction of the GP for the input~$\gpinput^{*}$ is a Gaussian distribution with posterior mean and variance given by
\begin{equation} \label{eq:one-step prediction mean and covariance}\\
\mu(\gpinput^{*}) = m(\gpinput^{*}) + {\vec k}\mat K^{-1} ({\vec \gpfun} - {\vec m})\,,\quad \Sigma(\gpinput^{*}) = k(\gpinput^{*},\gpinput^{*})-{\vec k}\mat K^{-1}{\vec k}^T\,,
\end{equation}
where $\mat K$ is the kernel matrix with $K_{ij}= k(\gpinput_i,\gpinput_j)$, $\vec m = [m(\gpinput_1),\ldots,m(\gpinput_l)]$ is the prior mean function, $\vec k =[k(\gpinput_1,\gpinput^{*}),\ldots,k(\gpinput_l,\gpinput^{*})]$ and $\vec \gpfun =[\gpfun(\gpinput_1),\ldots,\gpfun(\gpinput_l)]$. 
Thus, the GP provides both the expected value of the function at any arbitrary point~$\gpinput^{*}$ as well as a notion of the uncertainty of this estimate.
In this paper, we use GPs within BO (discussed in \sec{sec:BO}), to map policy parameters to predicted cost. In some of our simulations, we also use GPs to learn the unknown dynamics model $\dyn$ in \eq{eqn:dyn}, where, $\gpinput$ represents $(\state, \control)$ and $\gpfun$ represents $f$.
Central to our choice of employing GPs is their capability of explicitly modeling the uncertainty in the underlying function.
This uncertainty allows to account for the model-bias in the dynamics model, and to deal with the exploration/exploitation trade-off in a principled manner in BO.

\subsection{Bayesian Optimization (BO)} \label{sec:BO}
BO is a gradient-free optimization procedure that aims to find the global minimum of an unknown function~\cite{Kushner1964,Osborne2009,Shahriari2016}.
At each iteration, BO uses the past observations $\dataset=\{\gpinput_i,\gpfun(\gpinput_i)\}_{i=1}^l$ to model the objective function $\gpfun: \gpinput \rightarrow \gpfun(\gpinput)$, which is modeled using a GP. 
BO uses this model to determine the next informative sample location by optimizing the so-called acquisition function. 
Different acquisition functions are used in literature to trade off exploration and exploitation during the optimization process~\cite{Shahriari2016}.
In this work, we use the expected improvement (EI) acquisition function~\citep{movckus1975bayesian}.
Intuitively, EI selects the next parameter point where the expected improvement in performance is maximal. 
In this paper, we use BO as our MF method, i.e., we use BO to find the optimal policy parameters that minimize the cost function in \eq{eqn:cost_fcn} directly based on the observed cost on the system, as we will now detail in the next section. 


\section{Using Model-based Prior for Model-free RL}  
\label{sec:approach}
	\begin{algorithm}[t]
	\caption{MBMF Algorithm}
	\label{alg:MBMF}
	\begin{algorithmic}[1]
		\STATE \textbf{init}: Sample policy parameters $\params \sim N(0, I)$
		\STATE Apply sampled policies on the system and record resultant state-input trajectory and cost data
		\STATE Initialize $\dataset_1 \leftarrow \{(\state_k, \control_k), \state_{k+1}\}$;\quad $\dataset_2 \leftarrow \{\params, \cost(\params)\}$
		\STATE Train dynamics model $\dyn_{L}:{\state_k,\control_k} \rightarrow \state_{k+1}$ using $\dataset_1$
		\STATE Define $\cost_{L}(\cdot)$: Computed by evaluating the trajectory distribution corresponding to $\pol$ using Monte-Carlo on $\dyn_{L}$ and computing the expected cost in \eq{eqn:cost_fcn}
			\REPEAT 
			\STATE Train GP-based response surface $\respsurfNo:\params \rightarrow \cost_{\params}$ using $\dataset_2$ and $\cost_{L}(\params)$ as the prior mean
			\STATE Minimize the acquisition function $\acqfuncNo$: $\params^{'} = \minimize_\theta \, \acqfuncNo(\respsurfNo, \params)$ \label{line:acqFunc_opt}
			\STATE Evaluate $\params^{'}$ on the real system~$\dyn$
			\STATE Collect trajectory data $(\state_k, \control_k, \state_{k+1})$ and the observed cost $\cost(\params^{'})$
			\STATE Add $\{\params^{'},\cost(\params^{'})\}$ to $\dataset_2$ and trajectory data to $\dataset_1$
			\STATE \textbf{Every $F$ iterations:}
				\STATE ~~~Update the dynamics model $\dyn_{L}$ based on $\dataset_1$
				\STATE ~~~Redefine $\cost_{L}(\cdot)$ based on the updated GP dynamics
			\UNTIL{converged}
	\end{algorithmic}
	\end{algorithm}
	
We now present our novel approach to incorporating a MB prior in MF RL, which we term \textit{Model-Based Model-Free (MBMF)}. 
As with most MB approaches, our algorithm starts with training a forward dynamics model $\dyn_{L}$ from single-step state transition data $\dataset_1 := \{(\state_k, \control_k), \state_{k+1} \}$. 
This model can be linear or non-linear and can be learned in a variety of ways, e.g., using linear regression, GP regression, etc. 
Once the dynamics model is trained, for any given policy parameterization $\pol(\cdot, \params)$, we can predict the corresponding trajectory distribution by iteratively computing the distribution of states $\state_k := \dyn_{L}(\state_{k-1}, \control_{k-1})$ for $k = 1 \ldots T$.
Given the trajectory distribution, we compute the predicted distribution of the cost as a function of the policy parameters using \eq{eqn:cost_fcn}.
We denote the expected value of this predicted cost function as $\cost_{L}(\params)$

At the same time, similarly to BO, we train a GP-based response surface, $\respsurfNo(\params)$, that predicts $\cost(\params)$ given the measured tuples of $\dataset_2 = \{\params, \cost(\params)\}$.
Here, $\cost(\params)$ denotes the \textit{observed} cost corresponding to the policy $\pol(\cdot, \params)$ for the given horizon, as defined in \eq{eqn:cost_fcn}. 
However, unlike plain BO, we employ the prediction of the cost distribution from the dynamics model as the prior mean of the response surface\footnote{A more correct, but computationally harder approach would be to treat the full cost distribution as a prior for the response surface.}.
This modified response surface is then used to optimize the acquisition function $\acqfuncNo$ to compute the next policy parameters $\params^{'}$ to evaluate on the real system.
The policy $\pol(\cdot, \params^{'})$ is then rolled out on the actual system. 
The observed state-input trajectories and the realized cost data is next added to $\dataset_1$ and $\dataset_2$ respectively, and the entire process is repeated again. A summary of our algorithm is provided in \alg{alg:MBMF}.  

Intuitively, the learned dynamics model has the capability to estimate the cost corresponding to a particular policy; however, it suffers from the model bias which translates into a bias in the estimated cost. 
The BO response surface, on the other hand, can predict the true cost of a policy in the regime where it has observed the training samples, as it was trained directly on the observed performances.
However, it can have a huge uncertainty in the cost estimates in the unobserved regime. 
Incorporating the model-based cost estimates as the prior allows it to leverage the structure of the dynamics model to guide its exploration in this unobserved regime. 
Thus, using the model-based prior in BO leads to a sample-efficient exploration of the policy space, and at the same time overcomes the biases in the model-based cost estimates, leading to the optimal performance on the \textit{actual} system.        

Note that we collect trajectory data at each iteration so, in theory, we can update the dynamics model, and hence the response surface prior, at each iteration. 
However, it might be desirable to update the prior every $F$ iterations instead, as the dynamics model might change significantly between consecutive iterations, especially when the dataset $\dataset_1$ is small. 
We will demonstrate the effect of $F$ on the learning progress in \sec{sec:result}.

It should also be noted that algorithms like PILCO~\citep{Deisenroth2015} can be thought of as a special case of our approach, where the response surface consists exclusively of the prior mean provided from a GP-based dynamics model, without any consideration of the evidences (i.e., the measured costs).
In other words, PILCO does not take the dataset $\dataset_2$ into account.
Leveraging $\dataset_2$ allows the BO to learn an accurate response surface by accounting for the differences between the ``belief" cost based on the dynamics model and the actual cost observed on the system.

\begin{remark} \label{remark_meanfunc}
It is important to note that we do not explicitly compute the function $\cost_{L}(\params)$. The function is only computed for specific $\params$ that are queried by the optimization algorithm during the optimization of the acquisition function (Line \ref{line:acqFunc_opt} of \alg{alg:MBMF}).
\end{remark}

\begin{remark} \label{remark_generalmodels}
The proposed approach is agnostic to the function approximator used to learn the dynamics model; thus, different dynamics models, e.g., linear models, neural networks, GPs, Bayesian neural networks, etc. can easily be used in the proposed framework. 
\end{remark}


\section{Experimental Results}
\label{sec:result}
	In this section, we compare the performance of MBMF with a pure MB method, a pure MF method, as well as a combination of the two where the model is used to ``warm start" the MF method. 

\subsection{Experimental Setting}
	\paragraph{Task details}
	We apply the proposed approach as well as the baseline approaches on two different tasks. 
	In the first task, a 2D point mass is moving in the presence of obstacles.
	The setup of the task is shown in \fig{fig:trajs_task1}.
	The agent has no information about the position and the type of the obstacles (the Grey cylinders).
	The goal is to reach the goal position (the Green circle) from the starting position (the Red circle).  
	For the cost function, we penalize the squared distance from the goal position.

	In the second task, an under-actuated three degree-of-freedom (DoF) robotic arm (only two of the three joints can be controlled) is trying to push an object from one position to another. 
	The setup of the task is shown in \fig{fig:trajs_mujoco_task2_both}.
	The Red box represents the object which needs to be moved to the goal position, denoted by the Green box.
	As before, the squared distance from the goal position of the object is used as the cost function.

	These tasks pose challenging learning problem because they are under-observed, under-actuated, and have both contact and non-contact modes, which result in discontinuous dynamics. 

	\paragraph{Implementation details}
	For the GP regression, we use the GPy package~\cite{gpy2014}. 
	We use the Dividing Rectangles (DIRECT) algorithm~\cite{gablonsky2001modifications} for all policy searches in this paper.
	For simulating the tasks, we use OpenAI Gym~\cite{openAI_gym} environments and the Mujoco~\cite{todorov2012mujoco} physics engine. 
	In our experiments, we employ linear policies, but more complex policies can be easily incorporated as well. 

	\paragraph{Baselines details}
	For the MB method, we learn a dynamics model and use this dynamics model to perform policy search. 
	Given the dynamics model and the cost function, we learn a linear policy using DIRECT. 
	The resultant policy is then executed on the real system and the corresponding state and action trajectories, as well as the resultant cost are obtained. 
	The observed trajectories are then added to the training set, and the entire process is repeated again. 
	We denote this baseline as \textbf{MB} in our plots.

	For the MF method, we use BO to directly find the optimal policy parameters, and denote it as \textbf{MF} in plots. 
	In the final variant, we use the MB method above to optimize the policy for a given number of iterations, after which we switch to BO and continue the optimization. 
	The cost observations obtained during the executions of MB method were used to initialize the BO.
	We denote it as \textbf{MB+MF} in the plots.
	We will simulate this baseline for different switching points, which corresponds to the number of iterations after which we switch from MB to MF approach.
	We denote this number by \textbf{$K$} in our plots.
	Finally, we denote our approach as \textbf{MBMF} and also simulate it for multiple prior update frequencies $F$.  

\subsection{2D Point Mass}
The goal of this experiment is to demonstrate how leveraging the MB prior in the MF method can reduce the model-bias and yet maintains the data-efficiency.
We use a GP-based dynamics model for this simulation, where we learn a separate GP for every dimension of the state. 
We use Monte-Carlo simulation to find the trajectory distribution which is highly parallelizable and known to be very effective for GPs~\citep{kupcsik2014model}. 
Nevertheless, other schemes can be used to compute a good approximation of this distribution~\citep{Deisenroth2015}.
%
\begin{figure}[t]
\centering
\includegraphics[width=0.7\textwidth]{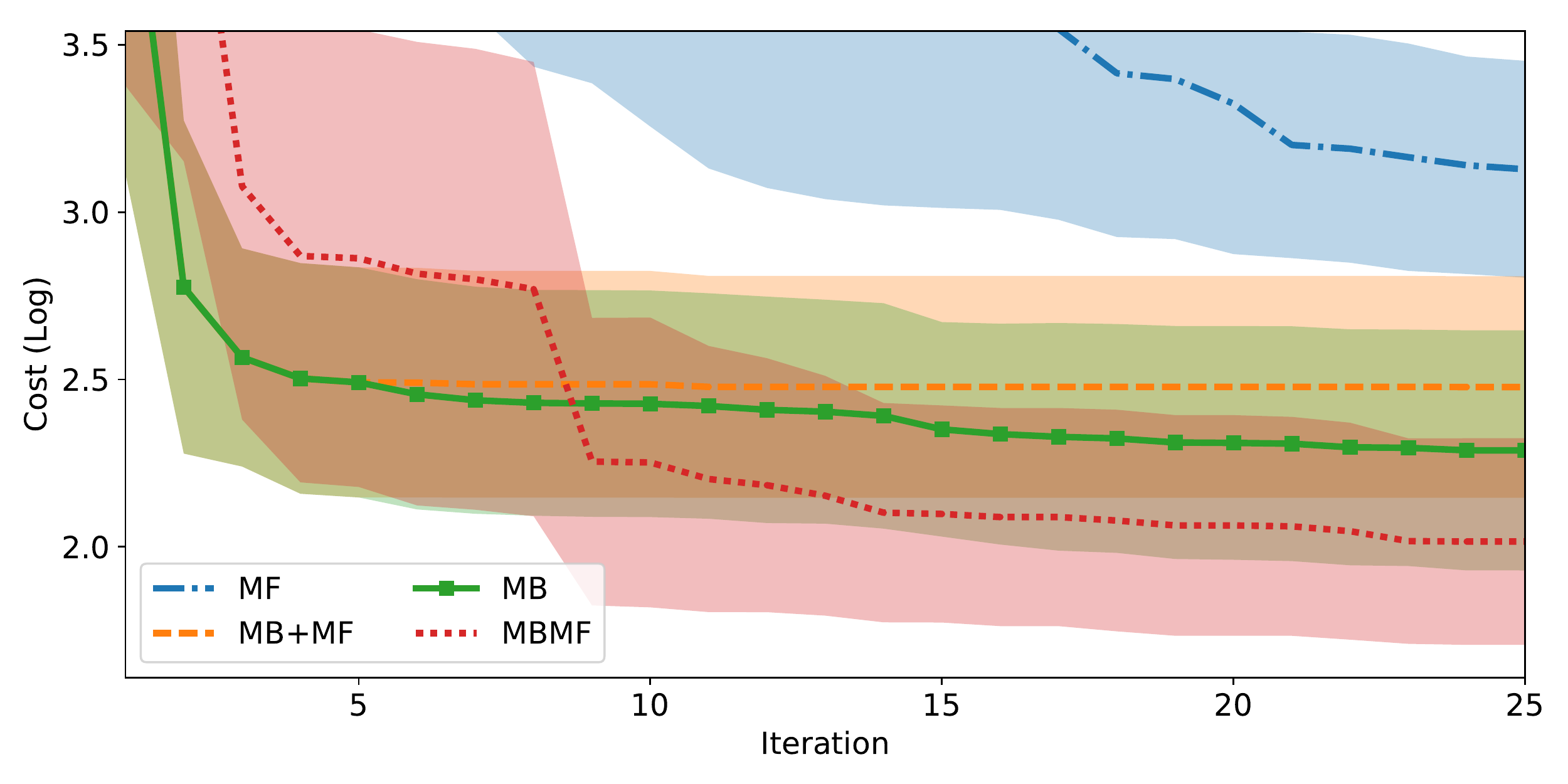}
\caption{The mean (curves) and the standard deviation (shaded regions) of the cost obtained for different approaches for the 2D point mass system.
  A pure MF approach is unable to perform well. 
  A pure MB approach continues to improve, but is outperformed by the MBMF, indicating the utility of blending MB and MF approaches.}
  \label{fig:results_task1}
\end{figure}
\begin{figure}[t]
\centering
\begin{subfigure}[b]{0.24\columnwidth}
\centering
  \includegraphics[width=\columnwidth]{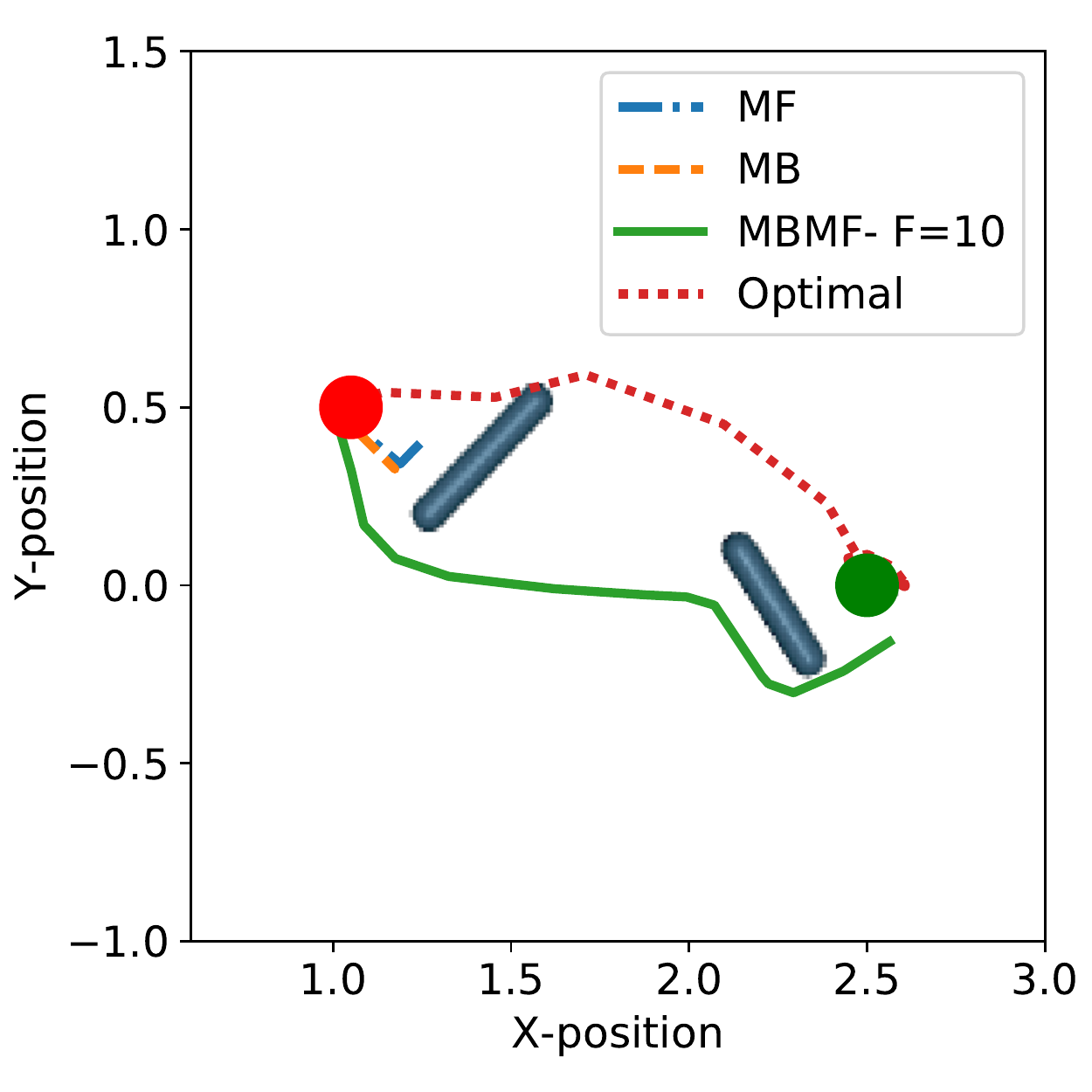}
  \subcaption{Trial 1}
  \label{fig:trial1_task1}
\end{subfigure}%
\hfill
\begin{subfigure}[b]{0.24\columnwidth}
\centering
  \includegraphics[width=\columnwidth]{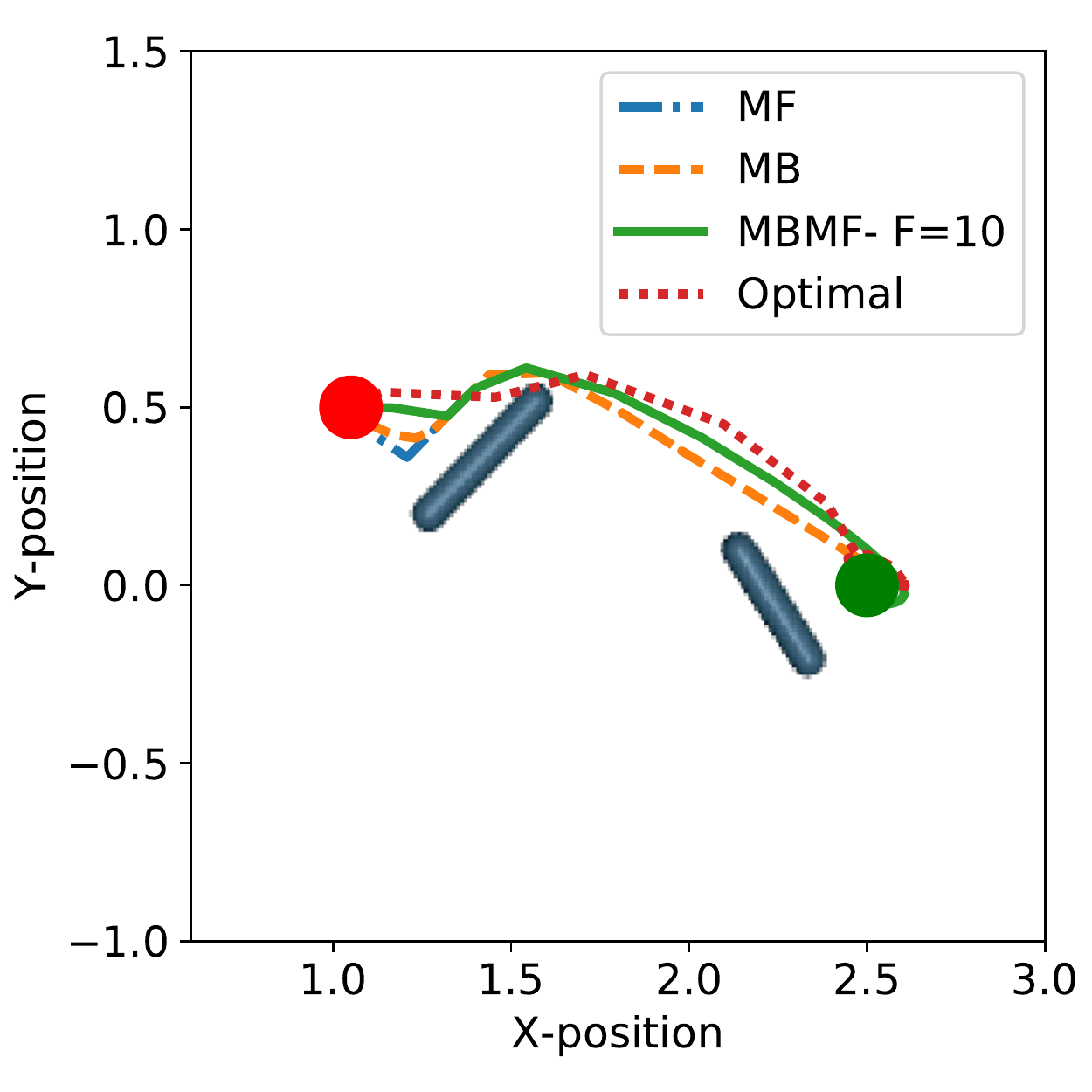}
  \subcaption{Trial 2}
  \label{fig:trial2_task1}
\end{subfigure}
\hfill
\begin{subfigure}[b]{0.24\columnwidth}
\centering
  \includegraphics[width=\columnwidth]{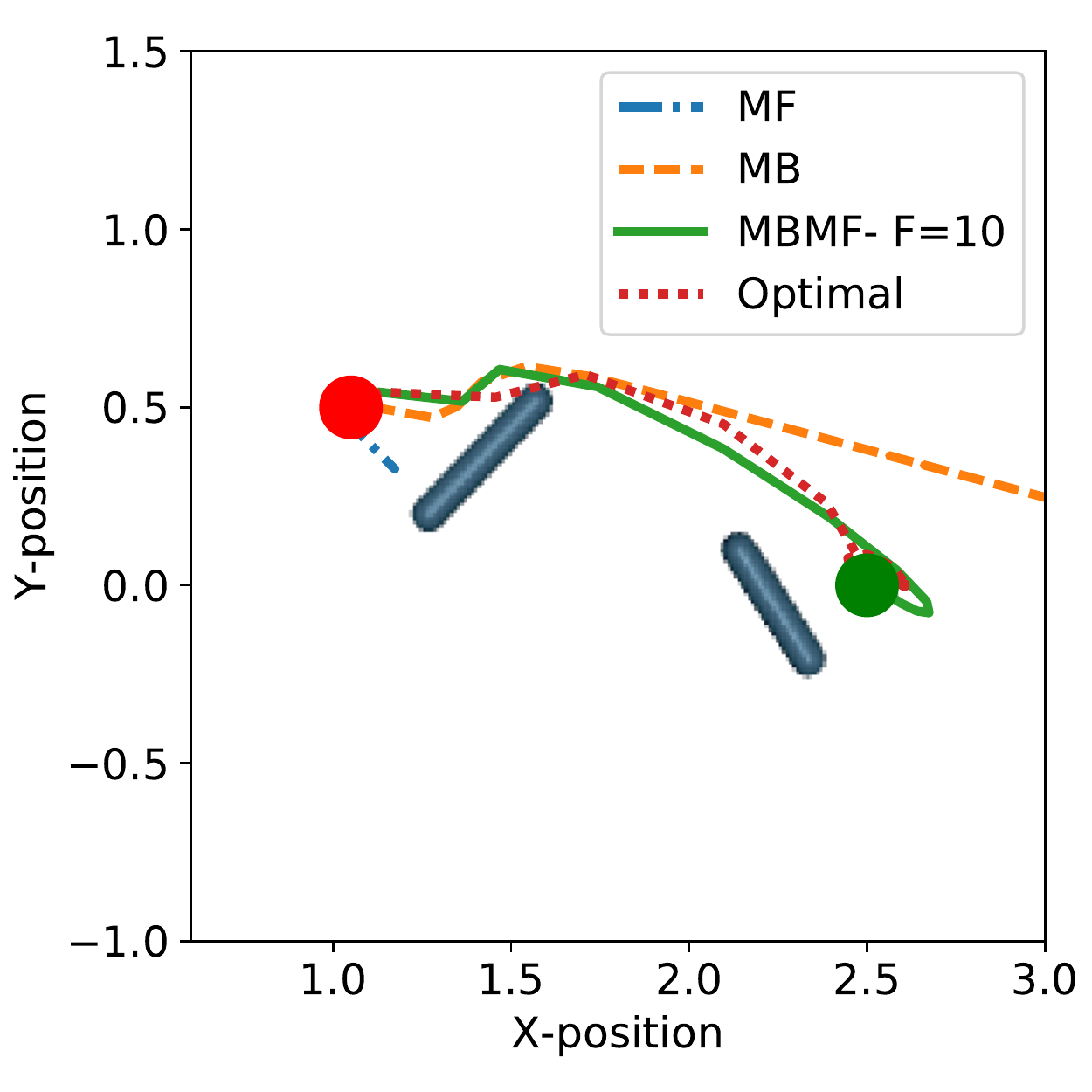}
  \subcaption{Trial 3}
  \label{fig:trial3_task1}
\end{subfigure}
\hfill
\begin{subfigure}[b]{0.24\columnwidth}
\centering
  \includegraphics[width=\columnwidth]{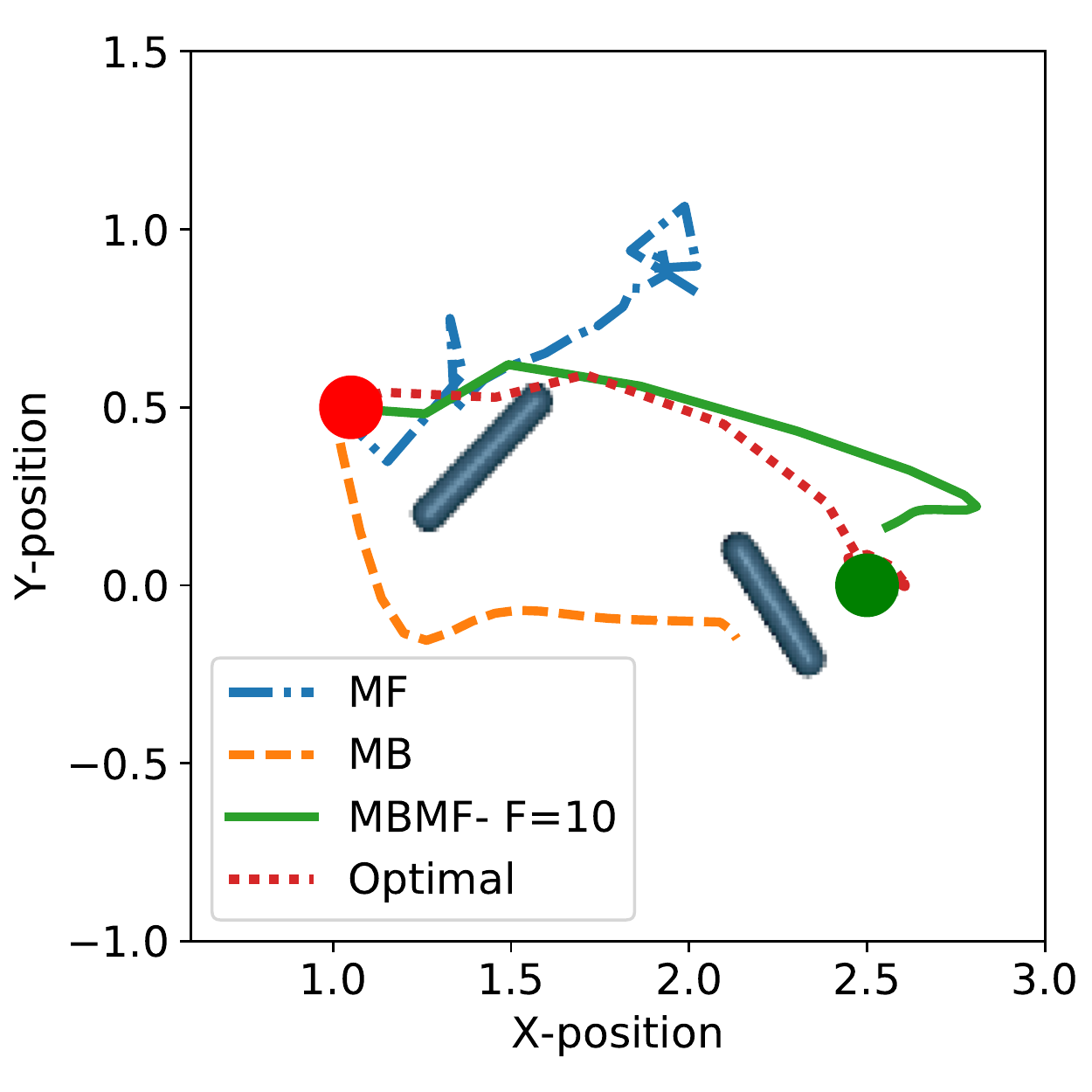}
  \subcaption{Trial 4}
  \label{fig:trial4_task1}
\end{subfigure}
\caption{Trajectories obtained via executing the learned controller for the point mass system after 25 iterations. 
Each trial corresponds to different initial data, but was same across all approaches.
The optimal trajectory requires the system to overcome the obstacles (the Grey cylinders) to reach from the initial position (the Red circle) to the goal position (the Green circle).
MB and MF approaches have different behavior across different trials and they often get stuck in the obstacles. 
MBMF, on the other hand, is able to learn how to overcome the obstacles and consistently reaches the goal position.  
}
\label{fig:trajs_task1}
\end{figure}

The optimal mean cost (curve) and the standard deviation (shaded area) obtained for different approaches (across thirty trials) as learning progresses are shown in \fig{fig:results_task1}, where each iteration corresponds to one execution on the real system. 
The MF approach (the dot-dashed Blue curve) improves as the learning progresses, but is still significantly outperformed by all other approaches, indicating the data-inefficiency of a pure MF approach.
The pure MB approach (the Green curve) continues to improve as learning progresses; however, it is outperformed by MBMF very early on. 
Interestingly, in this case, using MB method to warm start the MF method (with $K$=5) doesn't improve the performance, as evident from the dotted Orange curve, indicating that using the MB component to initialize the MF component may not be sufficient for the policy improvement. 
In contrast, using model information as a prior for the MF method (with $F$=10) outperforms the other approaches and is able to learn a good policy roughly within 15 iterations, indicating the utility of systematically incorporating the model information during policy exploration.
We note that MBMF also has a smaller variance compared to all the other baselines, indicating the consistency in its performance.  

We also simulated MB+MF and MBMF for different $K$ and $F$ respectively.
A naive switching from MB to MF fails to improve the policy even for different switching points, and thus are outperformed by the pure MB approach. 
The frequency $F$ at which the prior is updated in the MBMF approach, however, affects the learning process.
We found that switching the model prior too frequently or too slowly both might lead to a suboptimal performance. 
Switching too often makes MBMF too sensitive to the changes in the dynamics model, which can change significantly especially early-on in the learning, and can ``mis-guide" the policy exploration. 
On the other hand, switching too slowly may strip it of the full potential of the dynamics model.    
In this particular case, the optimal frequency turns out to be $F=10$ (i.e., the MB prior is updated every 10 iterations).
It might also be interesting to note that the MBMF approach is at least as good as the best baseline for all values of $F$. 
Nevertheless, systematically finding the optimal update frequency is an interesting future direction.
The mean and the standard deviation of the costs obtained by different approaches, as well as additional learning curves can be found in appendix \ref{sec:appendix1}).

We also plot the trajectories obtained by executing the learned controller on the actual system for the MB, MF and MBMF approaches in \fig{fig:trajs_task1}. 
The initial and the goal positions are denoted by the Red and the Green circles respectively. 
For comparison purposes, the globally optimal trajectory (the dotted Red curve) was also computed, using the actual system dynamics obtained through MuJoCo; however, the dynamics are unknown to any of the learning method. 
We plot the trajectories for different trials, which correspond to different (but same across all methods) initial data. 
As evident from the figure, the MBMF approach is consistently able to reach the goal state, whereas the MB and MF approaches fail to achieve a consistent good performance.
In particular, the optimal trajectory requires the system to overcome the obstacle next to the starting position. 
A pure MB approach is unable to consistently learn this behavior, potentially because it requires learning a discontinuous dynamics model. 
Consequently, it is often unable to reach the goal position and gets stuck in the obstacles (Figures \ref{fig:trial1_task1}, \ref{fig:trial4_task1}). 
Similarly, a pure MF approach is unable to learn to overcome the obstacles within 25 iterations.
MBMF approach, however, can take evidence into account and is able to overcome this challenge to reach the goal state \textit{consistently}, demonstrating its robustness to the training data, which is also evident from a lower variance in the performance of MBMF.

\subsection{Three DoF Robotic Arm}
\begin{figure}[t]
  \centering
  \includegraphics[width=0.7\textwidth]{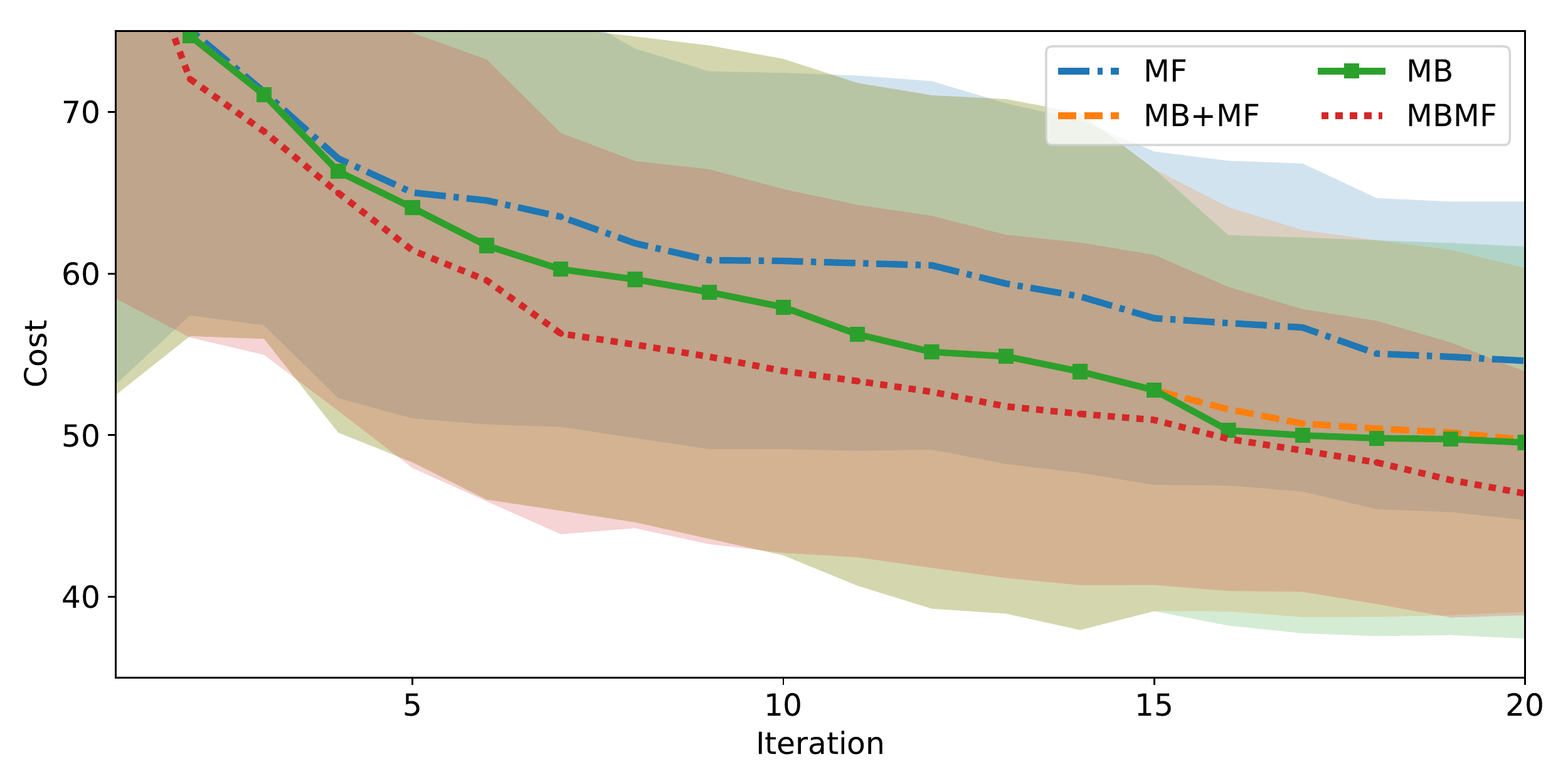}
  \caption{The mean (curves) and the standard deviation (shaded regions) of the cost obtained for different approaches for the three DoF robotic arm. 
MBMF leverages the advantages of both MB and MF approaches to design a better policy, indicating the data-efficiency of the MBMF approach, as well as its ability to overcome the model bias.
}
  \label{fig:results_task2}
\end{figure}

We again employ a GP-based dynamics model in this simulation.
As evident from \fig{fig:results_task2}, MBMF($F$=1) outperforms the other approaches and is continue to improve policy over iterations; however, due to the computational complexity of a GP-based dynamics model, we stop the learning process after 20 iterations. 
MB+MF approach($K$=15) continues to improve after switching from MB to MF; however, it is still outperformed by the pure MB approach.
We also note that MBMF has a significantly smaller variance compared to all other baselines, indicating that the MBMF approach is robust to the initial training data

We also simulate the MB+MF and MBMF approaches for different $K$s and $F$s. 
We only plot the curves corresponding to optimal $K$ and $F$ in \fig{fig:results_task2} for brevity reasons, but additional learning curves can an be found in appendix \ref{sec:appendix2}).
Interestingly, in this case, if the prior update frequency is too small ($F$ is large), then the MBMF lags behind the pure MB approach, as it is not fully leveraging the dynamics model information. 
However, if the right update frequency is chosen, then MBMF can leverage the advantages of both MB and MF approaches and outperforms the two.

\begin{figure}[t]
\centering
\begin{subfigure}[b]{0.99\columnwidth}
  \frame{\includegraphics[width=0.16\columnwidth]{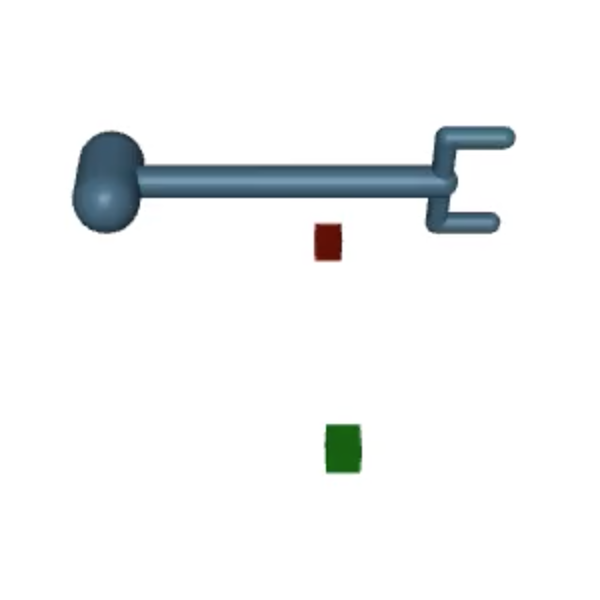}}\hfill
  \frame{\includegraphics[width=0.16\columnwidth]{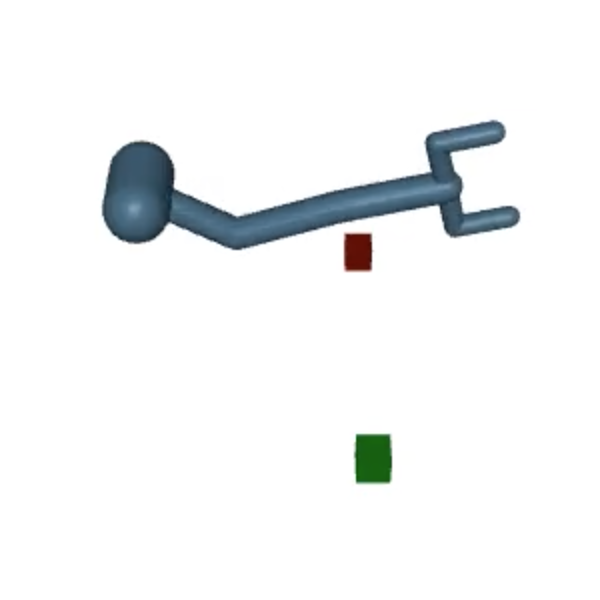}}\hfill
  \frame{\includegraphics[width=0.16\columnwidth]{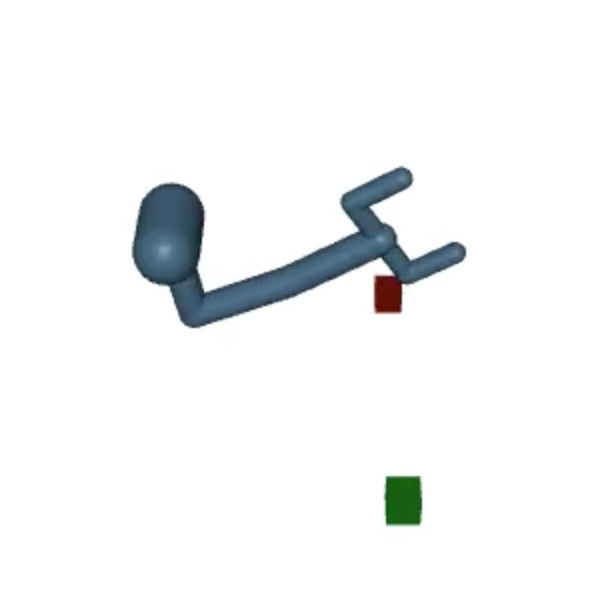}}\hfill
  \frame{\includegraphics[width=0.16\columnwidth]{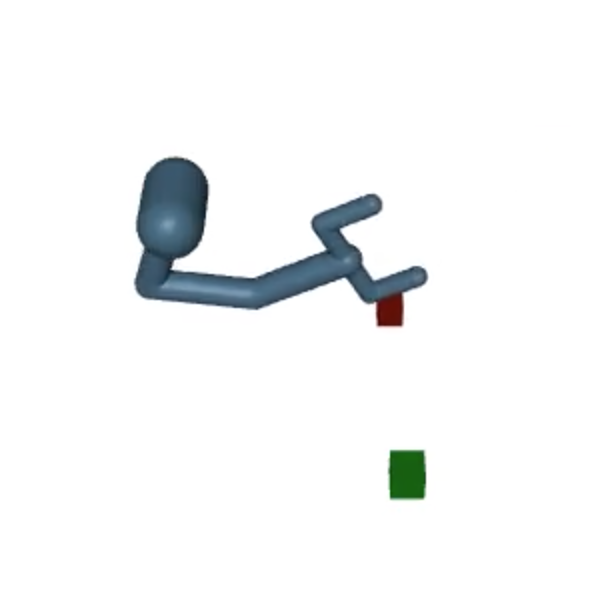}}\hfill
  \frame{\includegraphics[width=0.16\columnwidth]{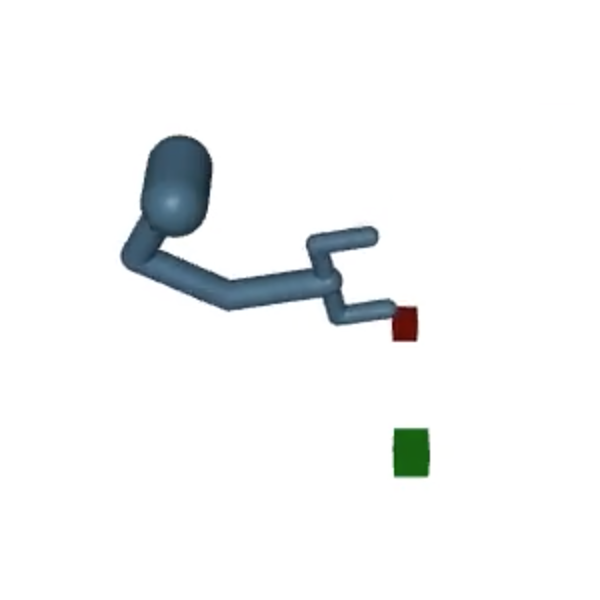}}\hfill
  \frame{\includegraphics[width=0.16\columnwidth]{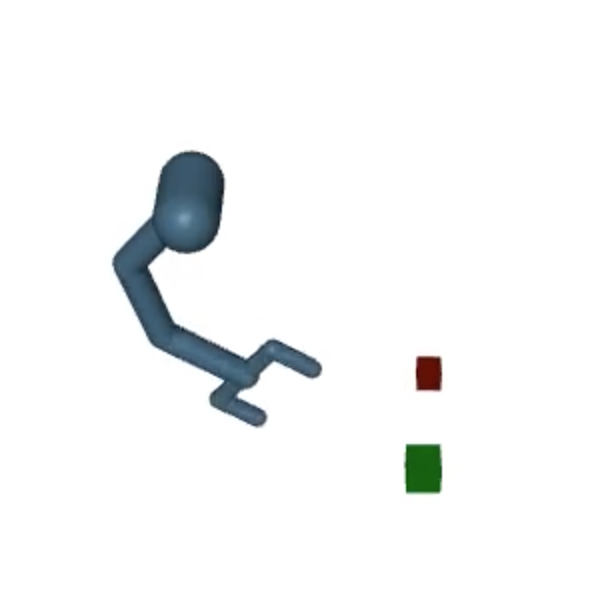}}
  \label{fig:trajs_mujoco_task2_1}
\caption{MBMF} 
\end{subfigure}
\\\vspace{5pt}
\begin{subfigure}[b]{0.99\columnwidth}
  \frame{\includegraphics[width=0.16\columnwidth]{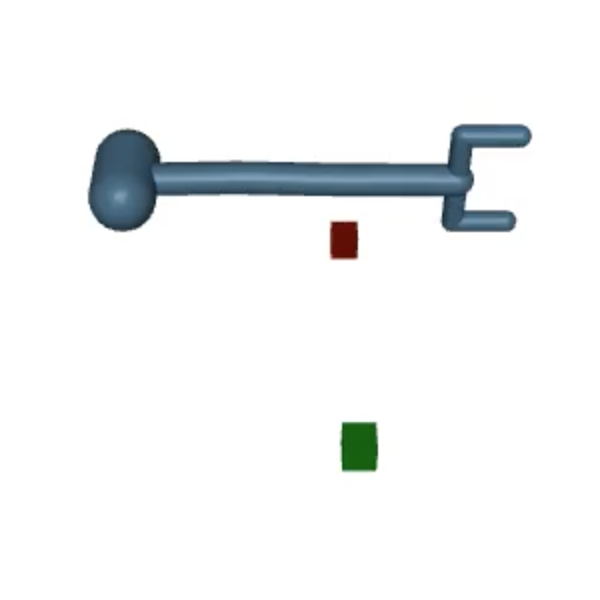}}\hfill
  \frame{\includegraphics[width=0.16\columnwidth]{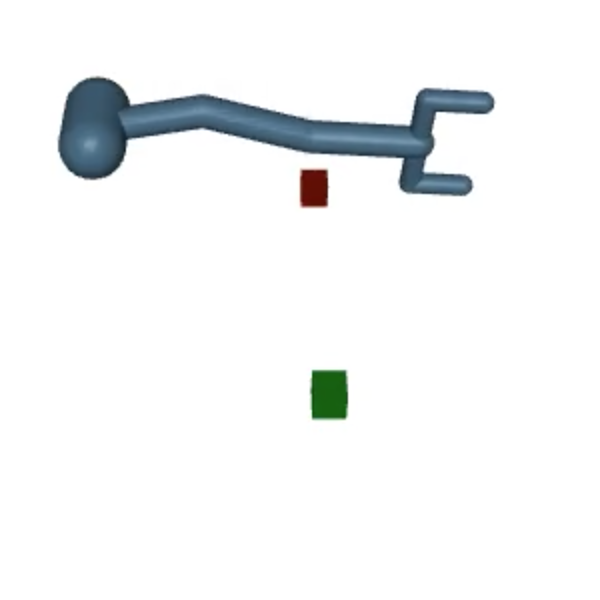}}\hfill
  \frame{\includegraphics[width=0.16\columnwidth]{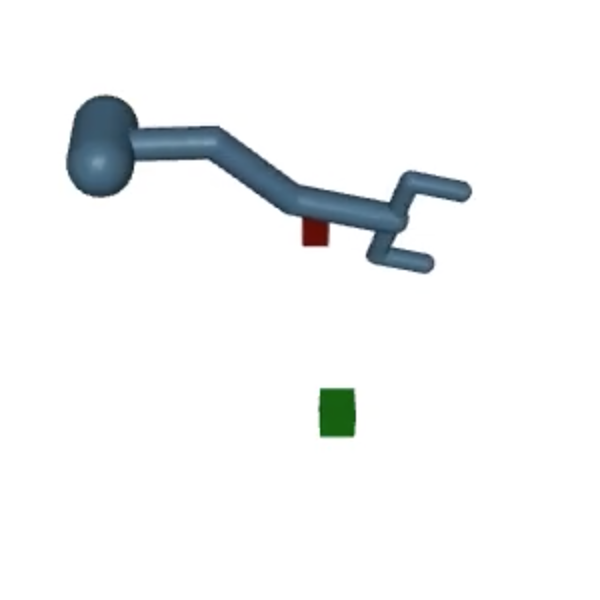}}\hfill
  \frame{\includegraphics[width=0.16\columnwidth]{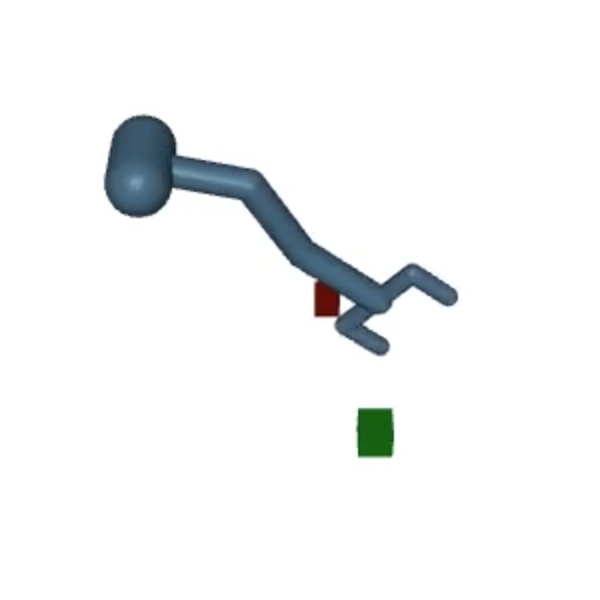}}\hfill
  \frame{\includegraphics[width=0.16\columnwidth]{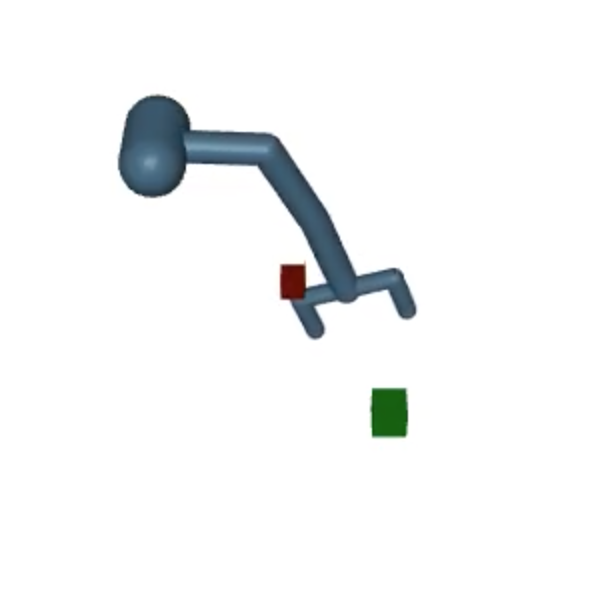}}\hfill
  \frame{\includegraphics[width=0.16\columnwidth]{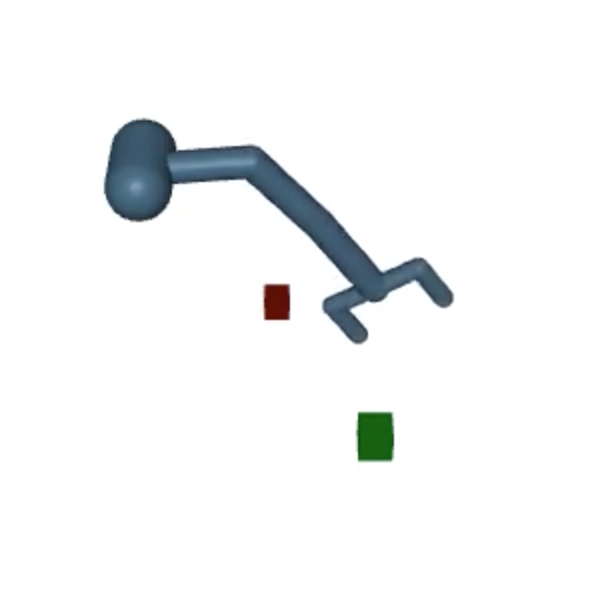}}
\label{fig:trajs_mujoco_task2_2}
\caption{MB}
\end{subfigure}
\caption{(a) Trajectory obtained via executing the learned controller for the MBMF approach. 
The Red box represents the object, which needs to be moved to the Green box.
MBMF is able to push the object fairly close to the goal position.
(b) Trajectory obtained via executing the learned controller for the pure model-based approach. 
A pure MB approach struggles with accomplishing this task, with the final position of the object end up being very far from the goal position.
}
\label{fig:trajs_mujoco_task2_both}
\end{figure}

The corresponding trajectory comparison between MB and MBMF approaches in Figure \ref{fig:trajs_mujoco_task2_both} also highlight the efficacy of MBMF in leveraging the advantages of both the MB and MF components to quickly learn the optimal policy. 
A pure MB approach struggles with learning to move the object vertically in a straight line, potentially due to the complexity of the dynamics given the contact-rich nature of the task. 
The MBMF approach, on the other hand, has the capability to trade-off the observed costs and the predicted cost. As a result, it has been able to move the object closer to the goal position within a small number of iterations (20 in this case).  
%


\section{Conclusion}
\label{sec:conclusion}
We propose MBMF, a novel probabilistic framework to combine model-based and model-free RL methods. 
This bridging is achieved by using the cost estimated by the model-based component as the prior for the model-free component.    
Our results show that the proposed approach can overcome the model bias and inaccuracies in the dynamics model, and yet retain the fast convergence rate of
model-based approaches. 
There are several interesting future directions that emerge out of this work.
First, it would be interesting to investigate how this approach performs on more complex tasks. 
Moreover, the prediction-time of Gaussian processes scales cubically with the number of training samples \cite{Rasmussen2006}, which makes the proposed approach prohibitive for the higher-dimensional systems or policies.
Exploring more scalable versions of the proposed approach is an interesting future direction.
Finally, a natural direction of research is the inclusion of other intermediate representations, such as value functions and trajectories, in the proposed approach.

\clearpage
\acknowledgments{This research is supported by NSF under the CPS Frontiers VehiCal project (1545126), by the UC-Philippine-California Advanced Research Institute under project IIID-2016-005, and by the ONR MURI Embedded Humans (N00014-16-1-2206).}


\bibliography{paper}  

\newpage
\section{Appendix}
\label{sec:appendix}
	\subsection{Point Mass System} \label{sec:appendix1}

\begin{figure}[H]
\centering
\includegraphics[width=0.5\columnwidth]{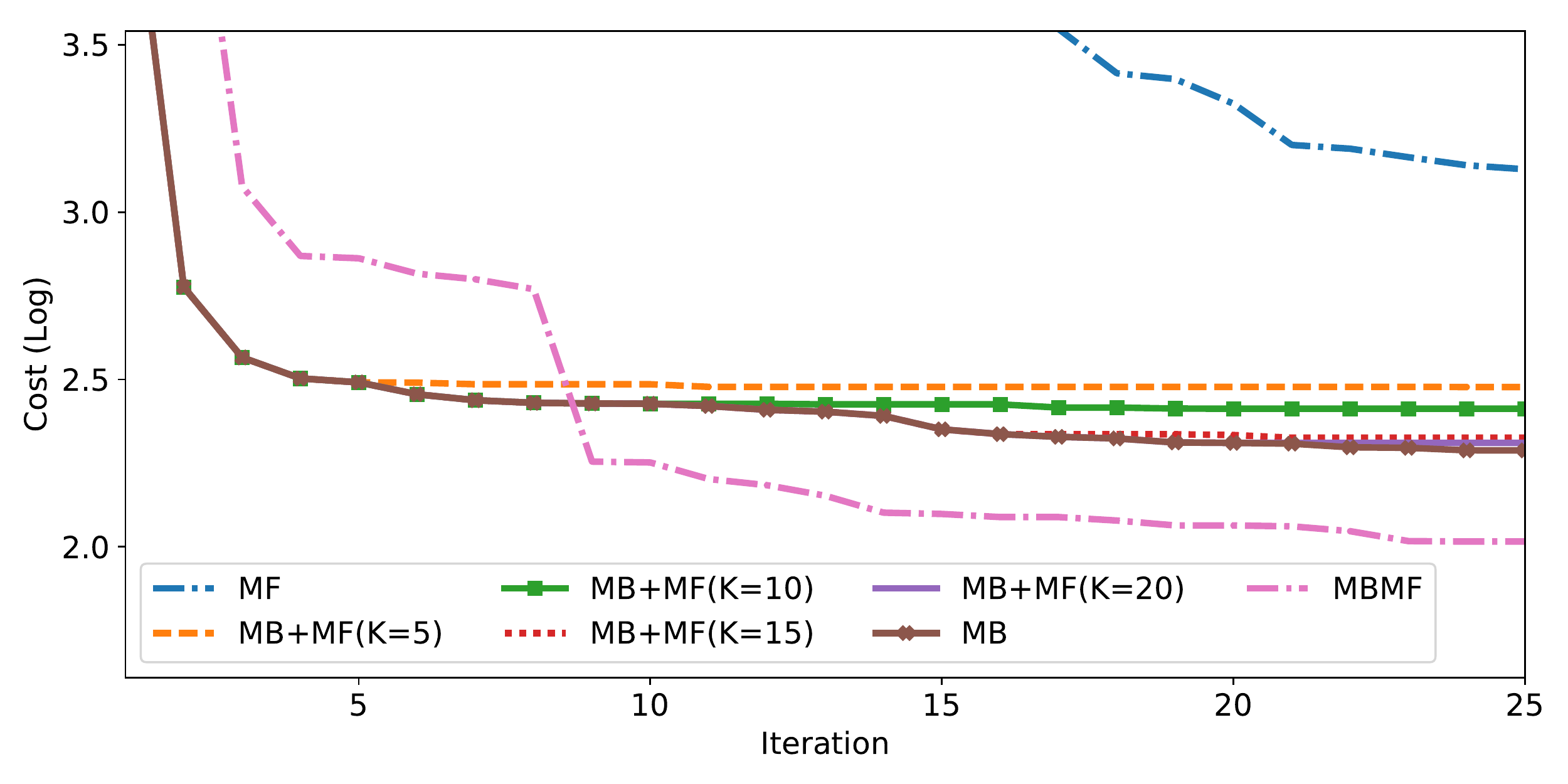}
\caption{The mean cost obtained for different switching points $K$ for the MB+MF approach for the 2D point mass system.
Switching from MB to MF results in a flat learning curve in this case, indicating that a naive switching between the two may not be sufficient for the policy improvement.}
  \label{fig:Kresults_task1}
\end{figure}
\begin{figure}[H]
\centering
\includegraphics[width=0.5\columnwidth]{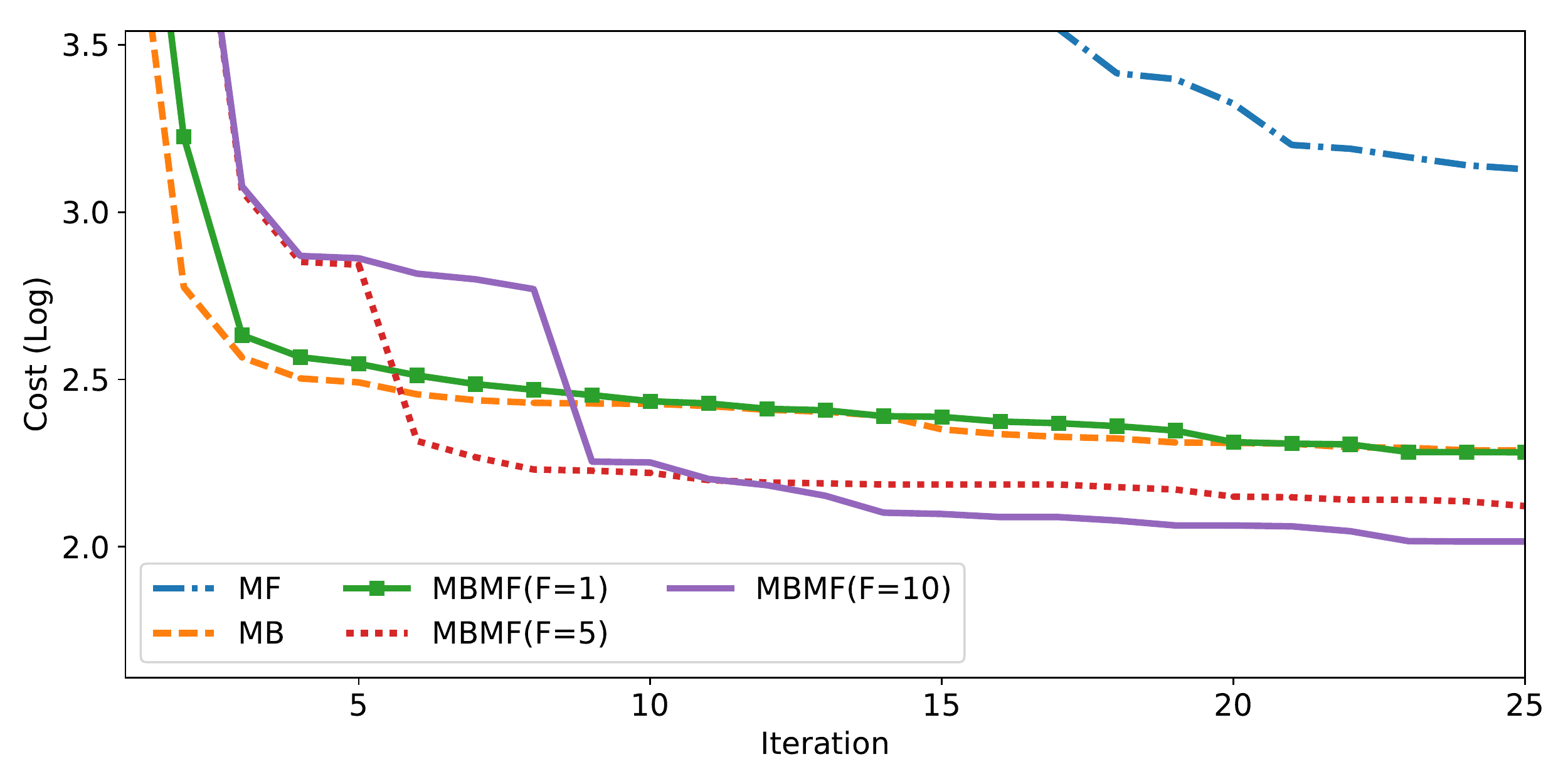}
\caption{The mean cost obtained for different prior update frequencies $F$ for the MBMF approach for the 2D point mass system.
The learning efficiency of MBMF depends on the choice of the prior update frequency.
Switching too often makes MBMF too sensitive to the changes in the dynamics model, which can ``mis-guide" the policy exploration. 
On the other hand, if the prior update frequency is too small ($F$ is large), then the MBMF lags behind the pure model-based approach, as it is not fully leveraging the dynamics model information. 
In this case, the optimal update frequency turns out to be $F=10$; however, MBMF is at least as good as the best baseline for all update frequencies.}
  \label{fig:Fresults_task1}
\end{figure}
\begin{table}[H]
\centering
    \begin{tabular}{| l | c |}
    \hline
    Approach &  Obtained Cost  \\ \hline
    Model-free (MF) & 22.83 $\pm$ 10.81  \\ \hline
    Model-based (MB) & 9.86 $\pm$ 4.10   \\ \hline
    MB+MF ($K=5$) & 11.91 $\pm$ 4.02  \\ \hline
    MB+MF ($K=10$) & 11.16 $\pm$ 3.87  \\ \hline
    MB+MF ($K=15$) & 10.24 $\pm$ 3.31  \\ \hline
    MB+MF ($K=20$) & 10.08 $\pm$ 4.02  \\ \hline
    MBMF ($F=1$) & \textbf{9.80 $\pm$ 2.62}  \\ \hline
    MBMF ($F=5$) & \textbf{8.34 $\pm$ 3.38}  \\ \hline
    MBMF ($F=10$) & \textbf{7.50 $\pm$ 2.93}  \\ 
    \hline
    \end{tabular}
    \vspace{1em}
    \caption{Point mass system. 
    The mean and the standard deviation of the cost obtained by executing the learned controller (after 25 iterations) on the actual system for different approaches. 
	The results are computed over 30 trials.	
	}
    \label{table:results_task1} 
 \end{table} 
\newpage
 
\subsection{Three DoF Robotic Arm} \label{sec:appendix2} 

\begin{figure}[H]
\centering
\includegraphics[width=0.5\columnwidth]{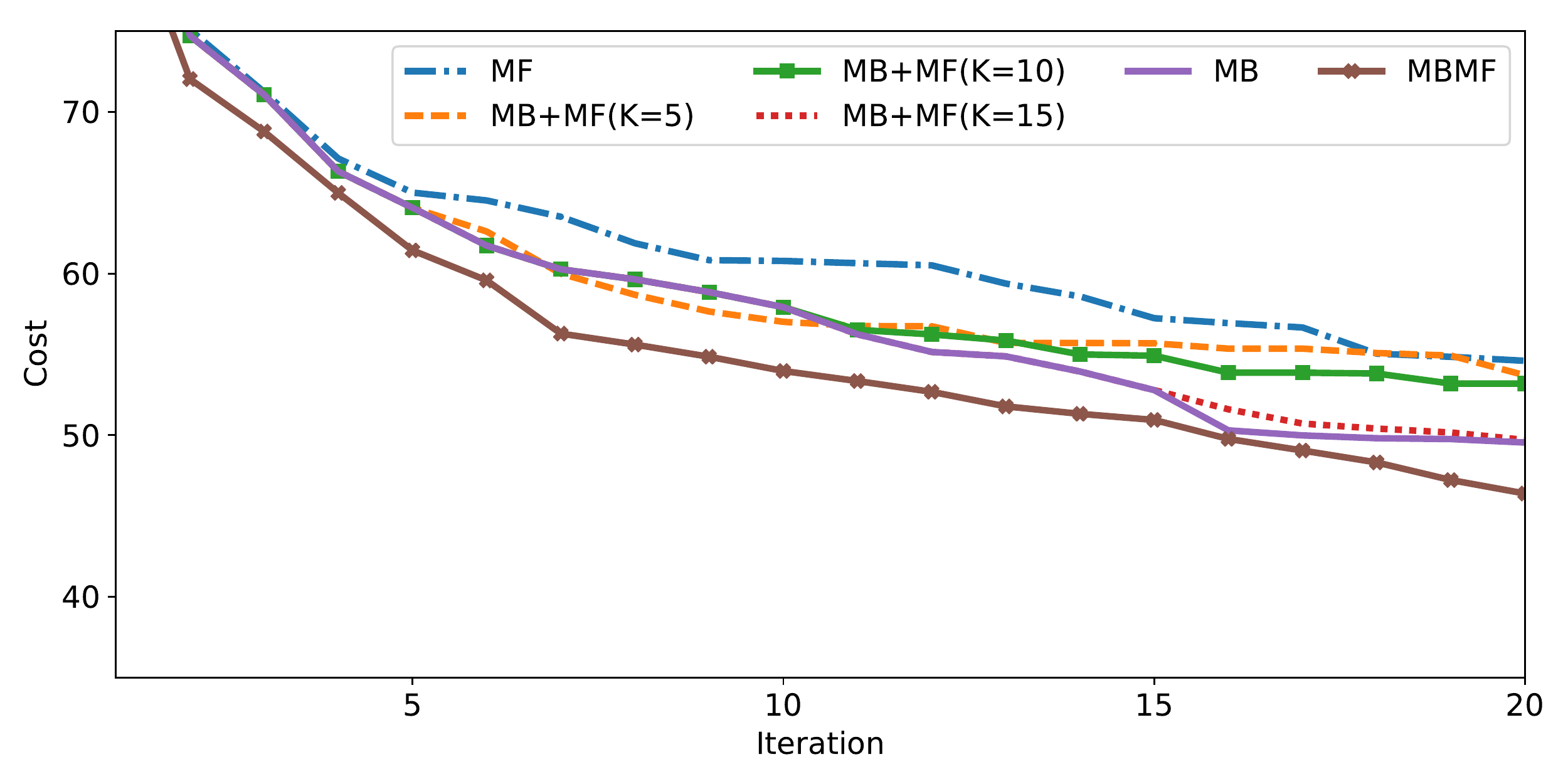}
\caption{The mean cost obtained for different switching points $K$ for the MB+MF approach for the robotic arm.
Switching from MB to MF results in a slower learning compared to a pure MB approach.}
  \label{fig:Kresults_task2}
\end{figure}
\begin{figure}[H]
\centering
\includegraphics[width=0.5\columnwidth]{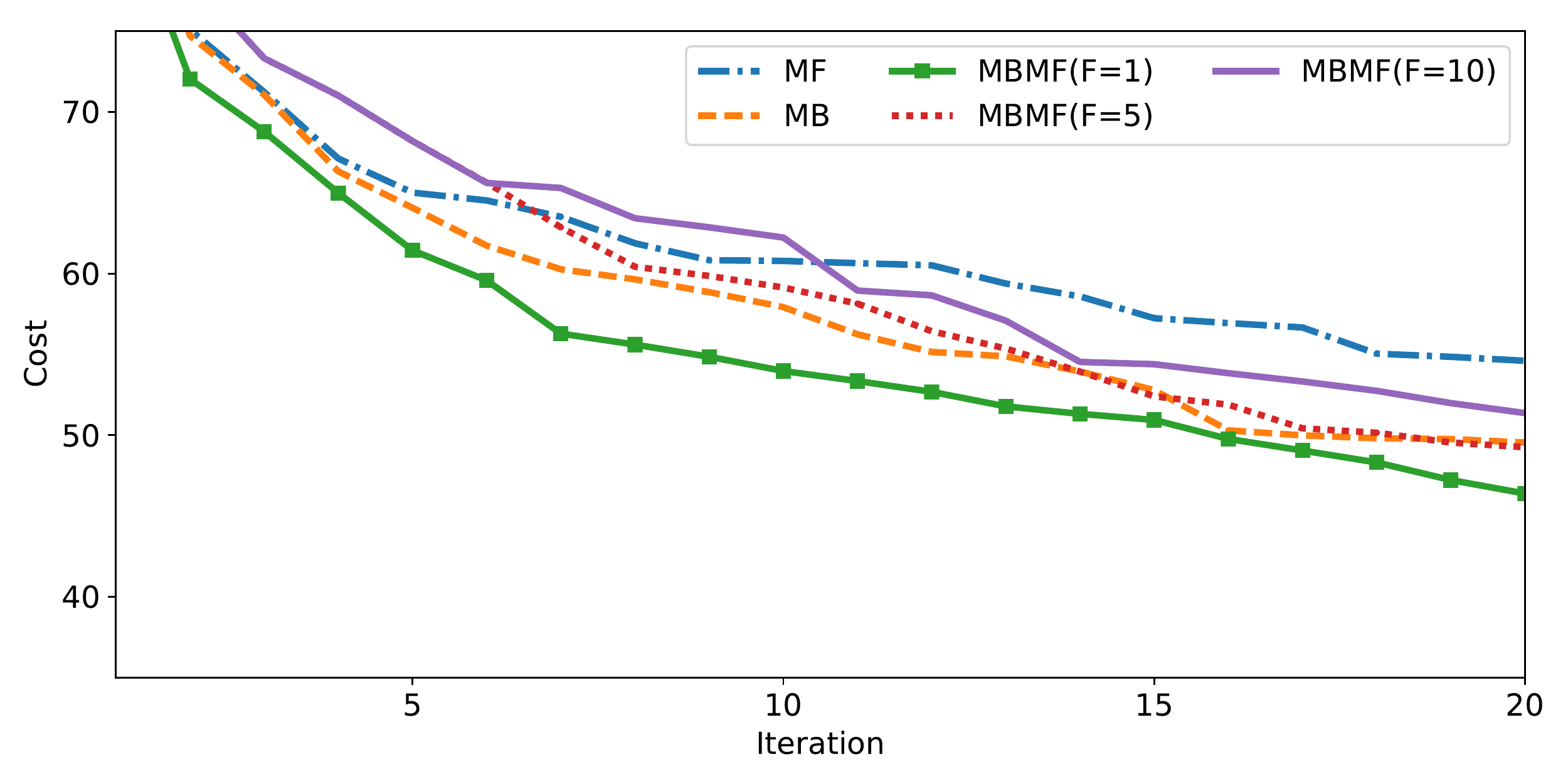}
\caption{The mean cost obtained for different prior update frequencies $F$ for the MBMF approach for the robotic arm.
If the prior update frequency is too small ($F$ is large), then the MBMF lags behind the pure MB approach in this case, as it is not fully leveraging the dynamics model information. 
In this case, the optimal update frequency turns out to be $F=1$.
Nevertheless, systematically finding the optimal prior update frequency is an important future direction.}
  \label{fig:Fresults_task2}
\end{figure}
\begin{table}[H]
\centering
    \begin{tabular}{| l | c |}
    \hline
    Approach &  Obtained Cost\\ \hline
    Model-free (MF) &       54.60 $\pm$ 9.85 \\ \hline
    Model-based (MB) &      49.54 $\pm$ 12.13 \\ \hline
    MB+MF ($K=5$) & 			53.70 $\pm$ 9.58 \\ \hline
    MB+MF ($K=10$) & 		53.19 $\pm$ 10.68 \\ \hline
    MB+MF ($K=15$) & 		49.70 $\pm$ 10.64  \\ \hline
    MBMF ($F=1$) &  		\textbf{46.38 $\pm$ 7.54} \\ \hline
    MBMF ($F=5$) & 		\textbf{49.25 $\pm$ 10.50} \\ \hline
    MBMF ($F=10$) & 		\textbf{51.36 $\pm$ 10.44} \\ 
    \hline
    \end{tabular}
    \vspace{1em}
    \caption{Three DoF robotic arm system. 
    The mean and the standard deviation of the cost obtained by executing the learned controller on the actual system for different approaches.
    The reported numbers are at the end of the iteration number 20 and are computed over 30 trials.	
	}
    \label{table:results_task2} 
 \end{table}  
%
 

\end{document}